\documentclass[9pt,twocolumn]{article}
\usepackage[a4paper, left=1.25cm, right=1.25cm, top=2.5cm, bottom=2.25cm]{geometry}
\usepackage[table,xcdraw]{xcolor}
\usepackage[utf8x]{inputenc}
\usepackage{graphicx}
\usepackage{placeins}
\usepackage{multirow}
\usepackage{romannum}
\usepackage{svg}
\usepackage{caption}
\usepackage{subcaption}
\usepackage{amsfonts}
\usepackage{amssymb}
\usepackage{amsmath}
\usepackage{cleveref}

\usepackage[
    backend=biber,
    style=nature,
    sorting=none,
    url=false
]{biblatex}  
\title{Latent Diffusion Models with Image-Derived Annotations for Enhanced AI-Assisted Cancer Diagnosis in Histopathology}

\author{
    Pedro Osorio
    \and
    Guillermo Jimenez-Perez
    \and
    Javier Montalt-Tordera
    \and
    Jens Hooge
    \and
    Guillem Duran-Ballester
    \and
    Shivam Singh
    \and
    Moritz Radbruch
    \and
    Ute Bach
    \and
    Sabrina Schroeder
    \and
    Krystyna Siudak
    \and
    Julia Vienenkoetter
    \and
    Bettina Lawrenz
    \and
    Sadegh Mohammadi$^1$
}

\date{Bayer AG \hspace{1em} \\ $^1$ \texttt{sadegh.mohammadi@bayer.com}
}
\makeatletter

\newcommand{\Rmnum}[1]{\expandafter\@slowromancap\romannumeral #1@}
\makeatother

\begin{document}

\pagenumbering{arabic}

\twocolumn[
\begin{@twocolumnfalse}
    \maketitle
    \begin{abstract}
    Artificial Intelligence (AI) based image analysis has an immense potential to support diagnostic histopathology, including cancer diagnostics. However,  developing supervised AI methods requires large-scale annotated datasets. A potentially powerful solution is to augment training data with synthetic data. Latent diffusion models, which can generate high-quality, diverse synthetic images, are promising. However, the most common implementations rely on detailed textual descriptions, which are not generally available in this domain. This work proposes a method that constructs structured textual prompts from automatically extracted image features. We experiment with the PCam dataset, composed of tissue patches only loosely annotated as healthy or cancerous. We show that including image-derived features in the prompt, as opposed to only healthy and cancerous labels, improves the Fréchet Inception Distance (FID) from 178.8 to 90.2. We also show that pathologists find it challenging to detect synthetic images, with a median sensitivity/specificity of 0.55/0.55. Finally, we show that synthetic data effectively trains AI models.
    \end{abstract}
    \vspace{1em}
\end{@twocolumnfalse}
]

\section{Introduction}

Histopathology, the gold standard for cancer diagnostics, involves the microscopic examination of tissue samples to discern manifestations of disease. These tissue samples, typically obtained through surgical resections or biopsies, are prepared and analyzed using hematoxylin and eosin (H\&E) staining protocols and can be digitized into gigapixel-sized whole-slide images (WSI). H\&E imaging offers insights into structural and morphological changes associated with various pathological conditions, including cancer. Due to their cost-effectiveness and accessibility, AI-based computer-assisted diagnosis (CAD) has already demonstrated immense potential by classifying diseases, detecting genetic alterations or quantifying lesions \cite{cersovsky2023towards, huss2020software, hohne2021detecting, sharma2023validation}.

However, applying this technology in the medical field, is challenging. First, collecting large enough datasets for model training is challenging due to disease rarity, to high acquisition costs and to reliance on low-availability technologies such as next-generation sequencing \cite{cersovsky2023towards, hohne2021detecting}. Second, histopathology-specific challenges arise, given the sheer size of the gigapixel-sized WSIs and the high variability of different staining and slide preparation techniques. This further complicates matters, as AI models often struggle to generalize, even with intricate model architectures \cite{vahadane2016structure, cersovsky2023towards, hohne2021detecting}. While traditional data augmentation offers a potential solution to address some of these issues, methods such as flipping and cropping often cannot adequately cover the full data distribution, whereas in-domain data augmentation techniques such as stain normalization can only slightly improve model generalization \cite{vahadane2016structure, hohne2021detecting}. This results in suboptimal improvements, as these transformations cannot effectively bridge the gaps introduced by missing data samples or address data imbalance or bias \cite{chen2022generative}.

Generative models are pivotal tools in the image synthesis field and have been used for many applications, ranging from bias mitigation, by augmenting an underrepresented class in a dataset, to privacy preservation, producing images that aren't derived from real subjects \cite{chen2021synthetic_private, dankar2021fake}. While some models, such as generative adversarial networks (GANs) have shown tremendous potential in generating diverse, distribution-wide and high-fidelity images, \cite{goodfellow_generative_2014, goodfellow_generative_2020, shorten2019surveyGANS, trabucco_effective_2023}, these often struggle with mode collapse and training instability, which hinders model training and demands meticulous hyperparameter adjustments \cite{thanh2020catastrophic}.

\begin{figure*}[t!]
    \centering
    \includegraphics[width=1\textwidth]{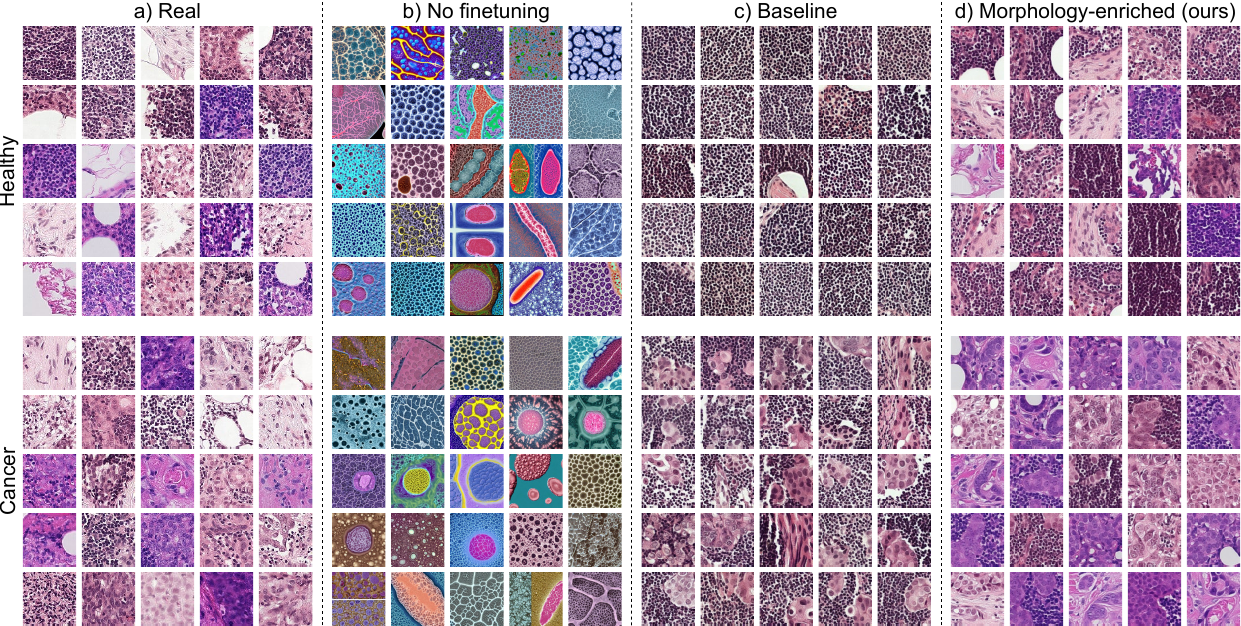}
    \caption{Randomly selected subset of 25 samples for: (a) the real dataset; (b) the synthetic set generated by Stable Diffusion (SD) out-of-the-box without fine-tuning; (c) the synthetic set generated by an SD model fine-tuned on histopathology data using a naïve prompt-building approach; and (d) the synthetic set generated by an SD model fine-tuned on histopathology data using our proposed prompt-building approach that leverages semantic information for improved generative diversity. Image grids are categorized per label (healthy, cancer).}
\label{fig:ft_vs_nonft}
\end{figure*}

Latent Diffusion Models (LDM), on their behalf, have recently gained attention for high-resolution text-conditioned image generation \cite{ho_denoising_2020, rombach_high-resolution_2022}. LDMs allow for generating images by gradually removing noise from a set of random perturbations. Several studies have shown that LDMs can generate high-fidelity images comparable to GANs \cite{azadi2022morphology}, presenting significant advancements over GANs in two principal domains: they assure consistent convergence, thereby addressing the training instability inherent to GANs, and intrinsically resist mode collapse, ensuring that the produced samples are diverse and accurately represent the underlying data distribution.

Given their benefits, research into diffusion models for histopathology image generation is growing steadily \cite{aversa2023diffinfinite, yellapragada2023pathldm, ye2023synthetic}. Aversa \textit{et al.} \cite{aversa2023diffinfinite} introduced an LDM-based sampling method that utilized annotated cellular macro-structure data from WSIs to produce arbitrarily large histological images, taking an important step towards WSI generation. However, their methodology relied not only on a diagnostic report at the WSI level but also on extensive, non-standard, and closed-source pixel-wise WSI annotations of 40 unique tissue types. In contrast, Yellapragada \textit{et al.} \cite{yellapragada2023pathldm} presented a method that conditions the image generation process with text report-based prompts at the WSI level and classifier-driven annotations at the patch level, yielding good generation metrics. Nevertheless, their strategy heavily relied on comprehensive WSI reports, which might not be representative for every medical problem. Finally, Ye \textit{et al.} \cite{ye2023synthetic} further minimized annotation requirements by firstly training an unconditional diffusion model (i.e., trained without any annotations) for then fine-tuning it with a smaller annotated dataset. Their approach, however, suffers from the aforementioned drawbacks of GANs, being prone to mode collapse \cite{thanh2020catastrophic}, and relies on the assumption of close similarity between pretraining and target data distributions. Such an extensive and diverse dataset can be costly to produce, potentially restricting the applicability of their approach.

This work presents a data-centric methodology to extract and synthesize textual morphology-rich descriptors from H\&E images, improving image generation as compared to not fine-tuning LDM models or only using labels as inputs (Figure \ref{fig:ft_vs_nonft}). Our approach harnesses LDM training while requiring only basic metadata such as 'cancerous' versus 'non-cancerous'. For this purpose, Patch Camelyon (PCam) \cite{veeling_rotation_2018, ehteshami_bejnordi_diagnostic_2017, kaggle_challenge}, a dataset composed of 96x96 px patches extracted from WSIs of lymph node sections, was selected. Each patch is loosely annotated with a binary label indicating presence or absence of metastatic tissue. Then, a pretrained image embedding model (DiNO \cite{caron_emerging_2021}), was leveraged to capture rich cellular configurations, overarching tissue structures, and distinctive pathological landmarks in the image patches, independently of the exhaustiveness of the dataset’s annotations. The outputs of the DiNO model were postprocessed using a K-Means clustering algorithm to agglomerate similar data points based on feature similarity, facilitating the generation of informative prompts/captions (e.g., \textit{``Histology image of \textbf{healthy} tissue, morphology type \textbf{five}''}). These prompts are in turn utilized as richer inputs for fine-tuning a pretrained LDM model (the open-source Stable Diffusion \cite{rombach_high-resolution_2022} model pretrained on natural images \cite{schuhmann2022laion}), thereby better representing the data distribution captured by model.

\begin{figure*}[t!]
    \centering
    \includegraphics[width=0.8\textwidth]{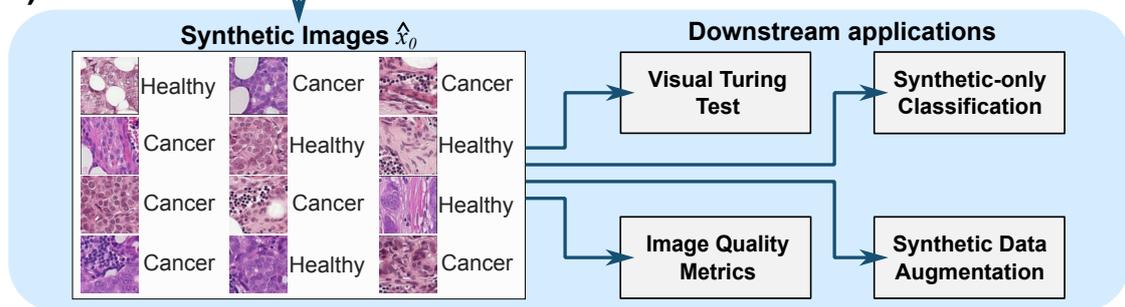}
    \caption{Overview of the pipeline proposed in this work. The inputs (a) consist in (image, label) pairs as curated from Patch Camelyon (PCam). The prompt-building pipeline (b) takes said inputs to construct a prompt (or caption) that describes each image in the input dataset. For this purpose, two approaches are followed: the baseline approach, in which only the label is used to generate a textual descriptor for the image; and the morphology-enriched approach, in which a frozen image embedder (DiNO \cite{caron_emerging_2021}) is used in combination with the patch's label to automatically extract semantic features from the image (clustered into 33 morphology types), to generate a morphology-rich prompt. After prompt-building, Stable Diffusion (c), an open-source Latent Diffusion Model (LDM), is trained using either of the prompt-building approaches from (b). Stable diffusion is based on a variational autoencoder (VAE) and a UNet. The VAE uses its encoder (E) to reduce the dimensionality of the input image into a latent ($z_0$), and can recover full-resolution images using its decoder (D). The VAE's latent ($z_0$) is used by the UNet, alongside the information in the prompt (via CLIP, a textual embedding model) to generate synthetic images. After model training, the performance of the fine-tuned stable diffusion model is evaluated on a series of downstream tasks (d). For this purpose, a large array of synthetic images are generated and tested using a visual Turing test, standard image quality metrics, and two classification approaches. The snowflake icon corresponds to a frozen model.}
\label{fig:overall_pipeline}
\end{figure*}

Next, the quality of the generated synthetic images is explored. Assessing the realism of synthetic images is a difficult endeavor that is highly debated in the scientific literature \cite{podell2023sdxl, kirstain2023pickapic}, so four downstream tasks were designed for testing this. Firstly, standard image quality metrics were computed to assess sample quality and coverage of the data distribution, namely the Fréchet Inception Distance (FID) \cite{heusel2017gans} and the improved precision and recall \cite{kynkaanniemi2019improved}. Secondly, a visual Turing test was performed for assessing whether pathologists could discern real from synthetic images. Finally, synthetic image quality was indirectly assessed via two distinct cancer detection subtasks: training a classifier using synthetic images only, to test how well the generative model replicates the discriminative features between cancerous and healthy tissue; and training classifiers where varying amounts of real and synthetic data are used for model training to evaluate the usability of the generated synthetic images for enhancing performance of CAD systems across multiple settings. Figure \ref{fig:overall_pipeline} delineates the main elements of the proposed methodology.

\begin{figure*}[t!]
    \centering
    \includegraphics[width=0.9\textwidth]{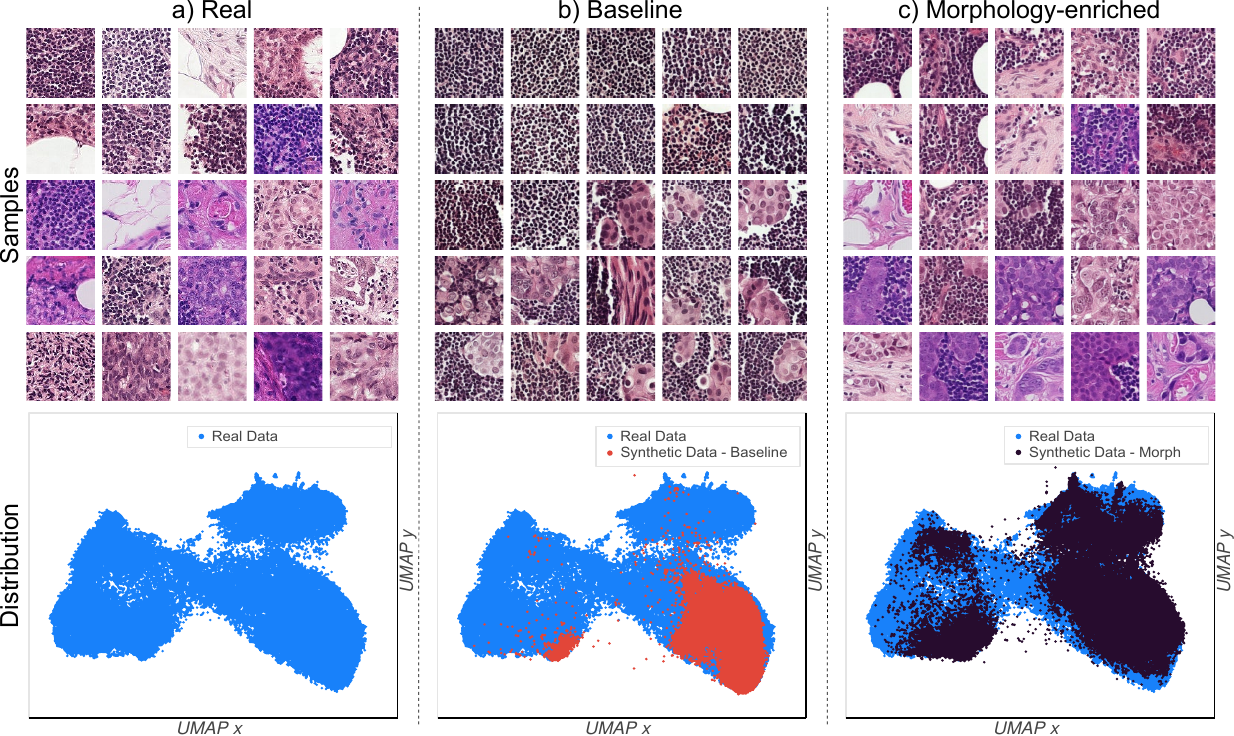}
\caption{Randomly selected subset of 25 samples for the real (a), baseline (b) and morphology-enriched (c) datasets. The rightmost column depicts the coverage comparison of the real data distribution between the two synthetic datasets. The manifold representation is generated based on Inception-v3 latents with a 2D UMAP transform.}
\label{fig:cancer_healthy_grids}
\end{figure*}

\section{Results}\label{sec:results}

\subsection{Data}\label{sec:results-data}

PatchCam (PCam) \cite{veeling_rotation_2018, ehteshami_bejnordi_diagnostic_2017, kaggle_challenge}, an open-source histopathology dataset comprising 327,680 patches extracted from 400 H\&E scans of lymph node sections from breast cancer patients, was selected for this work. This dataset includes no annotations other than a single binary class label per patch indicating the presence or absence of metastatic tissue in the center region, and predominantly comprises different cell morphologies: metastatic cancer cells, lymphocytes, and stroma. Section \ref{sec:methods-data} provides further details on the dataset characteristics. A total of 50k real image patches were selected after curation.

\subsection{Stable Diffusion can be fine-tuned to generate histological image patches}\label{sec:results-sd-baseline}

An initial finding was the inability of out-of-the-box SD models to generate histopathological images. This is likely due to the lack of digital pathology images in LAION, the large-scale dataset used for training vanilla SD \cite{schuhmann2022laion}. In fact, as illustrated in Figure \ref{fig:ft_vs_nonft}b, the images generated by this model without any additional training resemble more artistic interpretations of histopathological images, obviating the need for histopathology-specific fine-tuning.

To address this, an SD model was fine-tuned with in-domain medical data. This fine-tuning process further trains the vanilla SD model with histopathology image patches and corresponding prompts. In this experiment, the prompt for each of the patches were built by filling a template with the name of the corresponding class label: \textit{``Histology image of $\left\langle\textnormal{\textbf{LABEL}}\right\rangle$ tissue''}, where $\textnormal{\textbf{LABEL}}$ corresponds to either ``cancer'' or ``healthy''. After this fine-tuning step, the model trained with in-domain data was employed to generate class-conditional images from the healthy/cancerous categories. Figure \ref{fig:ft_vs_nonft}c depicts samples of the resulting synthetic dataset. 

Notably, the generated images are more realistic and show some level of control over their features (i.e., selectively generating images with or without cancer tissue) as compared to the non-fine-tuned counterpart, underlining the model's ability to represent complex biomedical concepts when captioned data is available. However, the synthetic images lack variability and seem to represent reduced color and morphological information, which could limit their utility for clinically relevant downstream tasks. To address this, further experiments varying the prompt-building process were performed, with the objective of maximizing information retained by the SD models from a dataset with limited annotations.

\subsection{Synthetic image quality can be improved using implicit image features}\label{sec:results-sd-proposed}

To improve upon the overall baseline image quality provided by the previous class conditional approach, the real data distribution must be better represented in the synthetic set. In accordance with the literature, \cite{brock2018large, dhariwal2021diffusion} we hypothesize that image diversity in the baseline approach is limited by the lack of conditioning variables in the dataset's metadata. This limits the descriptiveness of the constructed prompts, thereby hampering SD's ability to represent the full breadth of the real data distribution. To address this, we propose a morphology-enriched version of the baseline prompts that includes automatically extracted information from the real distribution. For this purpose, a novel prompt template was generated: \textit{``Histology image of $\left\langle\textnormal{\textbf{LABEL}}\right\rangle$ tissue, morphology type $\left\langle\textnormal{\textbf{INDEX}}\right\rangle$''}. Here, \textbf{INDEX} was computed by clustering features extracted from DiNO, a pretrained vision-transformer model \cite{caron_emerging_2021}. This index, therefore, represents particular sets of morphological features. Further detail can be found in Section \ref{sec:methods-pb}.

\begin{table}[t!]
    \centering
    \begin{tabular}{lccc}
        \hline
        \multicolumn{1}{l|}{\textbf{Prompting Strategy}} & \textbf{FID} $\downarrow$& \textbf{Precision} $\uparrow$& \textbf{Recall} $\uparrow$\\
        \hline
        \multicolumn{1}{l|}{\textbf{Baseline}} & 178.8& 0.065& \textbf{0.218}\\
        \multicolumn{1}{l|}{\textbf{Morphology-Enriched}} & \textbf{90.2} & \textbf{0.175} & 0.140\\
        \hline \\ 
    \end{tabular}
    \caption{Comparison of FID, improved Precision, and improved Recall metrics for the synthetic datasets generated via the baseline and morphology-enriched approach. Bold values indicate best performance, arrows indicate direction of improvement in the metric.}
    \label{tab:metric-comparison}
\end{table}

\begin{figure*}[t!]
    {
    \centering
    \includegraphics[width=0.75\textwidth]{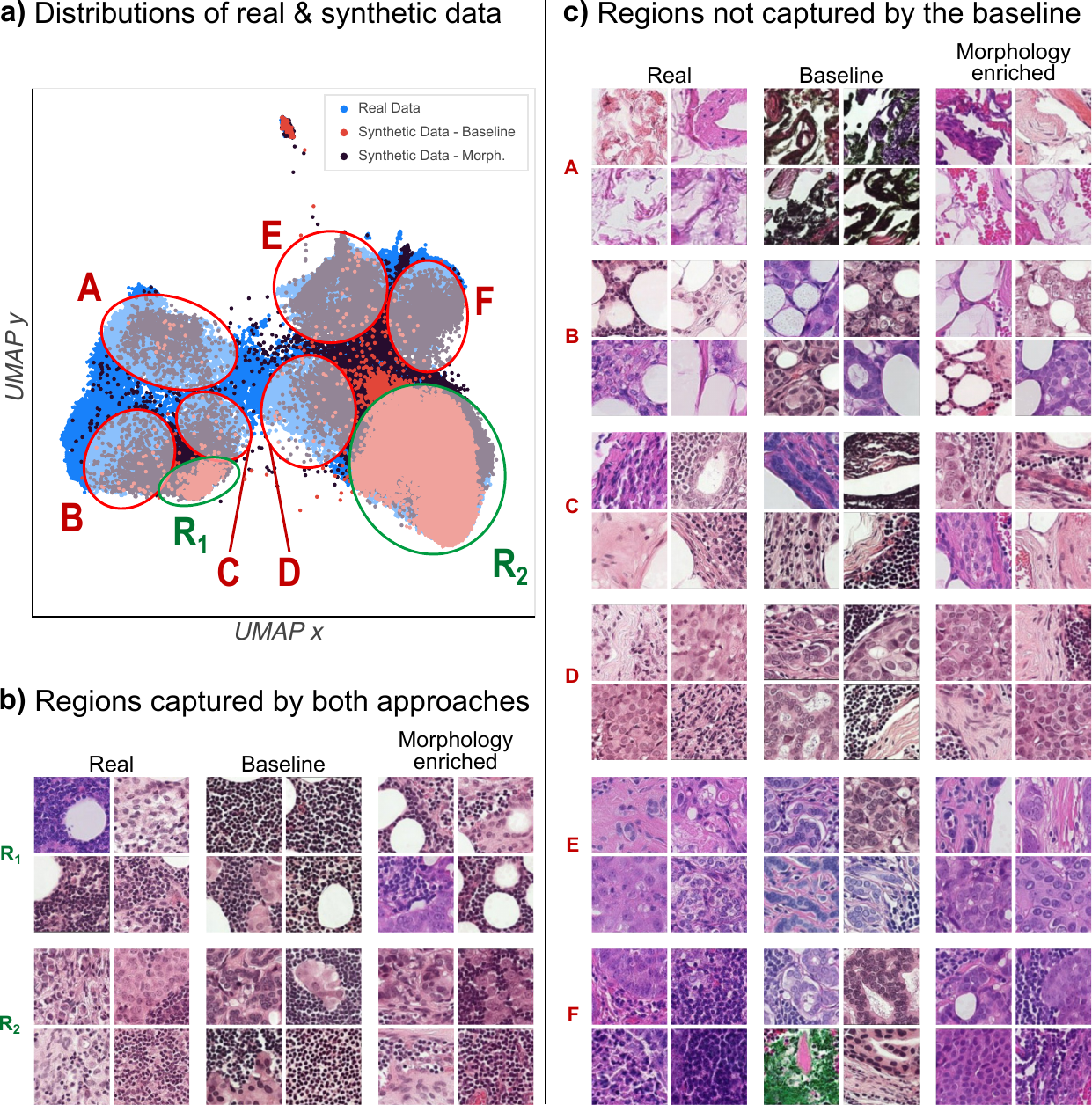}
    \caption{(a) UMAP Embedding Distributions: the real data distribution (blue) is better covered by the morphology-enriched prompt-building (black), as compared to the baseline prompts (red). Overlaid on the figure are regions R\textsubscript{1} and R\textsubscript{2} (in green), which are captured by both prompt-building approaches, and for which examples are selected in (b). Also overlaid are regions A through F (in red), which represents regions not captured by the baseline approach but well represented by the morphology-enriched prompt-building, with selected examples being depicted in (c). It is to note that the baseline approach is prone to visual outliers (e.g. green tincture in region F and darker images in region A). Zoom for detail.}
    \label{fig:coverage-with-samples-methods}
    }
\end{figure*}

Two main aspects were explored to address the image quality improvement over the baseline: closeness to real data distribution (fidelity) and degree of coverage of the real image distribution (diversity). First, a qualitative evaluation was conducted through the visual inspection of random subsets of real and synthetic data. The randomly selected subsets from Figure \ref{fig:ft_vs_nonft} and Figure \ref{fig:cancer_healthy_grids} depict samples from the real dataset and random subsets of images generated via each prompt-building approach. Regarding coverage of the real data distribution, the morphology-enriched method yields a synthetic dataset containing a much wider variety of images compared to the baseline approach, and which also better approximates the diversity in the real dataset. To further illustrate this, Figure \ref{fig:coverage-with-samples-methods} depicts the embedding distributions overlapped in the same plot. As can be seen, training SD using only the label as conditioning tends to produce images from solely two modes, covering significantly less of the real data distribution. Moreover, the baseline consistently struggles to sample from certain regions of the real data distribution (A through F). In these regions, the baseline is limited to a few artefact-prone images that fail to closely match the features of real images in those regions.

Additionally, standard image quality metrics were also computed. Table \ref{tab:metric-comparison} shows that using this morphology-enriched prompt-building leads to a significant improvement in image fidelity as quantified by the Fréchet Inception Distance (FID) \cite{heusel2017gans}. The baseline approach from earlier yielded 178.8 FID score while the morphology-enriched approach achieves an improved score of 90.2. Both synthetic image fidelity and the degree of coverage of the real image distribution can be quantitatively evaluated through the precision and recall \cite{kynkaanniemi2019improved, sajjadi2018assessing} metrics, respectively, which are also presented in Table \ref{tab:metric-comparison}. The morphology-enriched method improves precision at the cost of recall, which is discussed in depth in Section \ref{sec:discussion}.

These results indicate that fine-tuning an SD model leveraging additional conditional information extracted from the images themselves contributes to more realistic and diverse synthetic images, confirming the initial hypothesis.

\subsection{Implicit image features can be used to generate images that yield better cancer detection performance when training on only synthetic data}\label{sec:results-sd-proposed-cas}

\begin{table}[t!]
    \centering
    \begin{tabular}{l|c}
        \hline
        \textbf{Training Dataset} & \textbf{AUC} $\uparrow$\\
        \hline
        Real Data& 0.960\\
        \hline
        Synthetic Data - Baseline& 0.773\\
        Synthetic Data - Morphology-Enriched& \textbf{0.805}\\
        \hline
    \end{tabular}
    \caption{Comparison of test set AUC when training a ResNet-34 on real data and on synthetic data generated from each of the explored prompt-building approaches.}
    \label{tab:CAS-comparison}
\end{table}

Next, we trained a classifier on synthetic and real data respectively and assessed feature-based discriminative power for ``cancer'' and ``healthy'' image patches. The underlying assumption is that the better the generative model captures the true data distribution, the better the classification performance in models trained on the resulting synthetic data \cite{ravuri2019classification}.

Table \ref{tab:CAS-comparison} presents the classification results from this analysis. As expected, training solely on real data still outperformed training on either of the synthetic datasets. However, training on the synthetic dataset generated through our prompt-building approach yields a higher AUC score when compared to the baseline, reducing the gap towards the real data by 0.032 AUC. This can be attributed to the more fine-grained control that the SD model trained on the morphology-enriched prompts has over the image space. Consequently, the morphology-enriched method seems to allow an improved capacity to reproduce the meaningful features responsible for the distinction between the two classes in the dataset, which further supports the hypothesis raised in the previous section.

\subsection{Distinguishing synthetic images from real was challenging even to expert pathologists}

To further evaluate synthetic image quality, we designed and executed a Visual Turing Test to assess the plausibility of the generated images as reviewed by five board-certified veterinary pathologists. This analysis revealed that the readers were not able to reliably distinguish the real from the synthetic images generated via the morphology-enriched method under the conditions of the test. For this purpose, a standardized set comprising 40 images (20 real and 20 synthetic), was presented to each reader. Each reader was tasked with classifying each image as either real or synthetic, while conveying their level of certainty through the qualifier ``definitely'' or ``maybe''. In addition, readers could provide comments along with their answers, which were recorded for later analysis. Further details about this experiment can be found in \ref{sec:methods-visual_turing}.

\begin{figure*}[t!]
    \centering
    \includegraphics[width=1\textwidth]{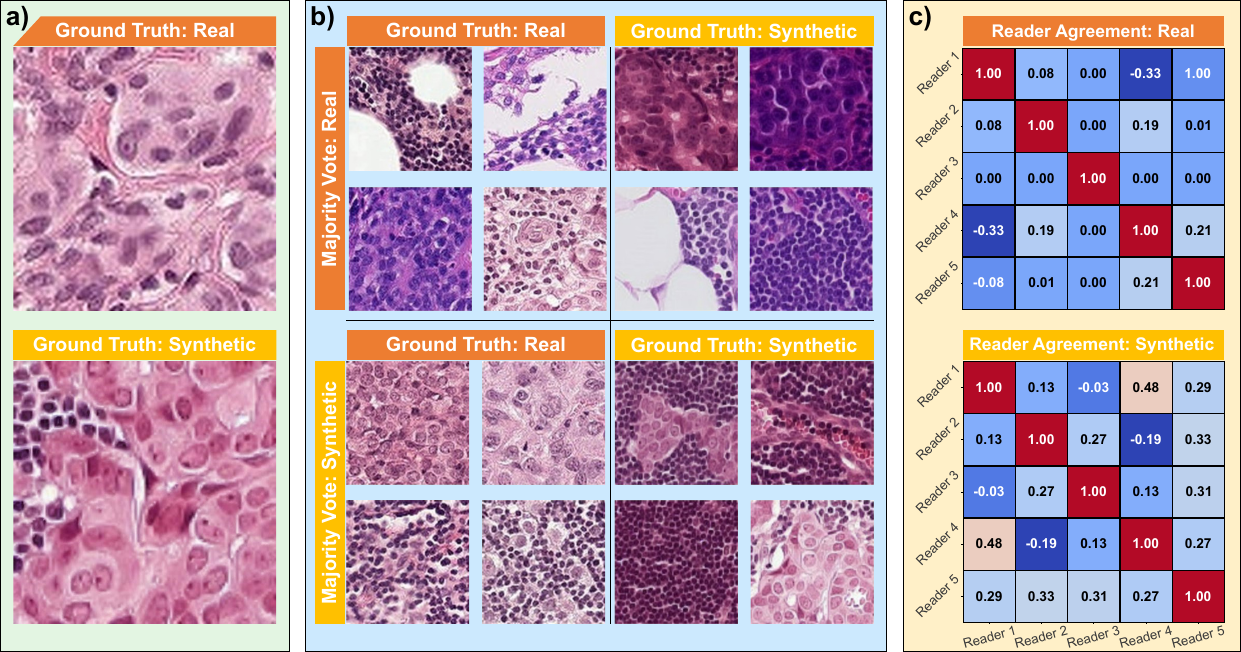}
    \caption{Reader agreement analysis results. (a) Real and synthetic images highlighting the resampling artifacts visible on real but not on synthetic images. Synthetic images showed a smoother visual appearance. (b) Selected true positive and false positive examples with the highest inter-reader agreement. Slightly higher inter-reader agreement was found when the ground truth was synthetic, irrespective of whether the majority reader decision was a true positive or not. (c) Inter-reader reliability based on pairwise Cohen's Kappa coefficients for readers' label decisions on real (top) and synthetic images (bottom). The overall agreement for each of these scenarios is reported as the mean ($\mu$) and standard deviation ($\sigma$) of the kappa coefficients in the off-diagonal.}
    \label{fig:turing_reader_uncertainty_agreement_artifact}
\end{figure*}

The results are presented in Table \ref{tab:turing_reader_perf} and show that the accuracy for detecting synthetic images ranged from 0.15 to 0.65, with a median value of 0.55. Reader sensitivity ranged from 0.3 to 0.7, with a median value of 0.55, while specificity ranged from 0.0 to 0.6, with a median value of 0.55. No statistically significant discriminative power was found for four out of five readers. The best-performing reader achieved a sensitivity of 0.7 and a specificity of 0.6 (p-value = 0.081). Interestingly, reader 3 obtained statistically significant accuracy results, although their criteria were inverted: they classified most real images as synthetic and vice versa. Closer visual inspection revealed the presence of artifacts in real images, likely due to upscaling of the PCam images. In fact, synthetic images do not display this artifact and appear smoother (see Figure \ref{fig:turing_reader_uncertainty_agreement_artifact}a). Indeed, while revising the study images, one of the readers was able to exploit this observation to classify the images with a near-perfect accuracy of 0.95. This artifact may have consciously or subconsciously been picked up by reader 3 as well. The other four readers dismissed it as a non-relevant image artifact as opposed to relevant cellular morphology.

\begin{table*}[t!]
    \centering
    \begin{tabular}{l|cccccccc}
        \hline
        \textbf{Reader} & \textbf{Accuracy} & \textbf{Sensitivity} & \textbf{Specificity} & \textbf{NPV} & \textbf{PPV} & \textbf{p-value} & \textbf{Confidence}\\
        \hline
        1 & 0.65 & 0.70 & 0.60 & 0.67 & 0.64 & 0.081 & 0.025\\
        2 & 0.50 & 0.45 & 0.55 & 0.50 & 0.50 & 1.000 & 0.000\\
        3 & 0.15 & 0.30 & 0.00 & 0.23 & 0.00 & $<$0.001 & 0.000\\
        4 & 0.55 & 0.55 & 0.55 & 0.55 & 0.55 & 0.636 & 0.000\\
        5 & 0.57 & 0.70 & 0.45 & 0.56 & 0.60 & 0.430 & 0.000\\
        \hline
    \end{tabular}
    \caption{Visual Turing test results. We report the accuracy, sensitivity, specificity, positive predictive value (PPV) and negative predictive value (NPV) for each reader in the detection of synthetic images (positive class = “synthetic”). We also report the p-value for the null hypothesis that readers cannot discriminate between real and synthetic images. The confidence of each reader in their responses was measured by the proportion of instances where the ``definitely'' was used to describe their choice.}
    \label{tab:turing_reader_perf}
\end{table*}

As depicted in Table \ref{tab:turing_reader_perf}, readers overwhelmingly preferred the low-confidence qualifier ``maybe'' over the high-confidence ``definitely'', with only one answer across all readers labeled with the latter. The reader's comments help explain this finding, with a focus on poor image quality and blurriness (of both real and synthetic images) as a limiting factor in their ability to decide whether the image was real or synthetic. Lead times (time to answer) were also measured and interpreted as a proxy for reader confidence. No significant difference in lead time was found between real and synthetic image. (Full lead time analysis can be found as supplementary material).

Finally, the inter-reader agreement was analyzed via the Cohen’s Kappa statistic \cite{mchugh2012interrater}. Overall agreement was very close to chance ($\mu=0.07$, $\sigma=0.12$). 
The agreement among readers when the images were synthetic ($\mu=0.20$, $\sigma=0.18$) was larger than when the images were real ($\mu=0.01$, $\sigma=0.14$). However, none of these values exceed the reliability criterion ($\kappa=0.21$, see Section \ref{sec:methods-visual_turing}). Nevertheless, inter-reader agreements (depicted in Figure \ref{fig:turing_reader_uncertainty_agreement_artifact}c) did, for some reader pairs, exceed the criterion, particularly for synthetic images. It cannot be excluded that this higher agreement in synthetic images, alongside with their slightly higher positive predictive values (Table \ref{tab:turing_reader_perf}), could suggest the presence of outliers among synthetic examples that were correctly identified by most readers. Figure \ref{fig:turing_reader_uncertainty_agreement_artifact}b provides examples of images with the highest agreement.

\subsection{Augmenting a small real training set with synthetic data improves performance of computer-aided diagnosis systems}
\label{sec:results-synth_aug}

A classification task was designed to assess the performance improvement when training on multiple proportions of real and synthetic data, with the objective of evaluating the utility of the morphology-enriched generation pipeline. The methodology for comparison is further described in Section \ref{sec:methods-synth_aug}.

Figure \ref{fig:multiple_dr} aggregates the test set AUC values across 10 cross-validation folds for various settings. Section \ref{sec:methods-synth_aug} further describes the experimental setup used. As anticipated, the test set AUC exhibits a positive correlation with the quantity of real data used to train the classifier. Additionally, for data regimes up to 500 real training samples, there is a distinct performance improvement as the amount of added synthetic data increases. Notably, when having only 25 real samples and using and augmentation ratio of 300\% the median performance rises from 0.775 AUC to 0.831 AUC (0.056 increase) surpassing the performance when training on double the amount of real data (0.827 when training on 50 real samples). Also, using an augmentation ratio of 300\% when training with 50 real samples leads to a median performance of 0.865 AUC, which is a 0.038 increase from when just using real data (0.827) and is at the level of when training on double the amount of real data, i.e. 100 real samples (0.872 AUC). Similarly, training on 500 real samples and augmentation ratio of 300\% allows a median performance of 0.911 AUC which is a 0.007 increase from the fully real baseline and only slightly less than when training on 1000 real samples (0.916 AUC). Nevertheless, the above-mentioned trend plateaus when the initial amount of real data is already substantial and the classifier performance trained solely on real data already exceeds 0.9 AUC. Particularly, when the initial amount of real data is 10000 samples, then adding synthetic data can actually lead to a slight decrease in median test performance (0.946 on only real data to 0.943 when the ratio is 50\% and 200\%).

\section{Discussion}\label{sec:discussion}

This paper presents a novel prompt-building approach to enable high-quality histopathological image generation using an open-source text-to-image diffusion model (Stable Diffusion) despite the lack of detailed textual descriptions. Although pretrained latent diffusion models are excellent tools for synthetic image generation, they require fine-tuning on target images paired with detailed textual descriptions to bridge the domain gap between the pretraining and target distributions. However, such textual information is frequently lacking in the context of medical imaging, particularly in digital pathology. As such, for addressing the low informativeness of the labels contained in PCam (cancer/non-cancer), the proposed morphology-enriched prompt-building method leverages a pretrained vision-transformer model, which converts the image dataset to a lower-dimensional embedding space, and a K-means model, which discretizes the latter forming a set of clusters with shared image features. This two-step approach complements the existing metadata in the dataset by extracting enhanced representations of the input patches without any external information, to complete the scarce annotations and better capture the diversity of the input dataset.

\begin{figure*}[t!]
    \centering
    \includegraphics[width=\textwidth]{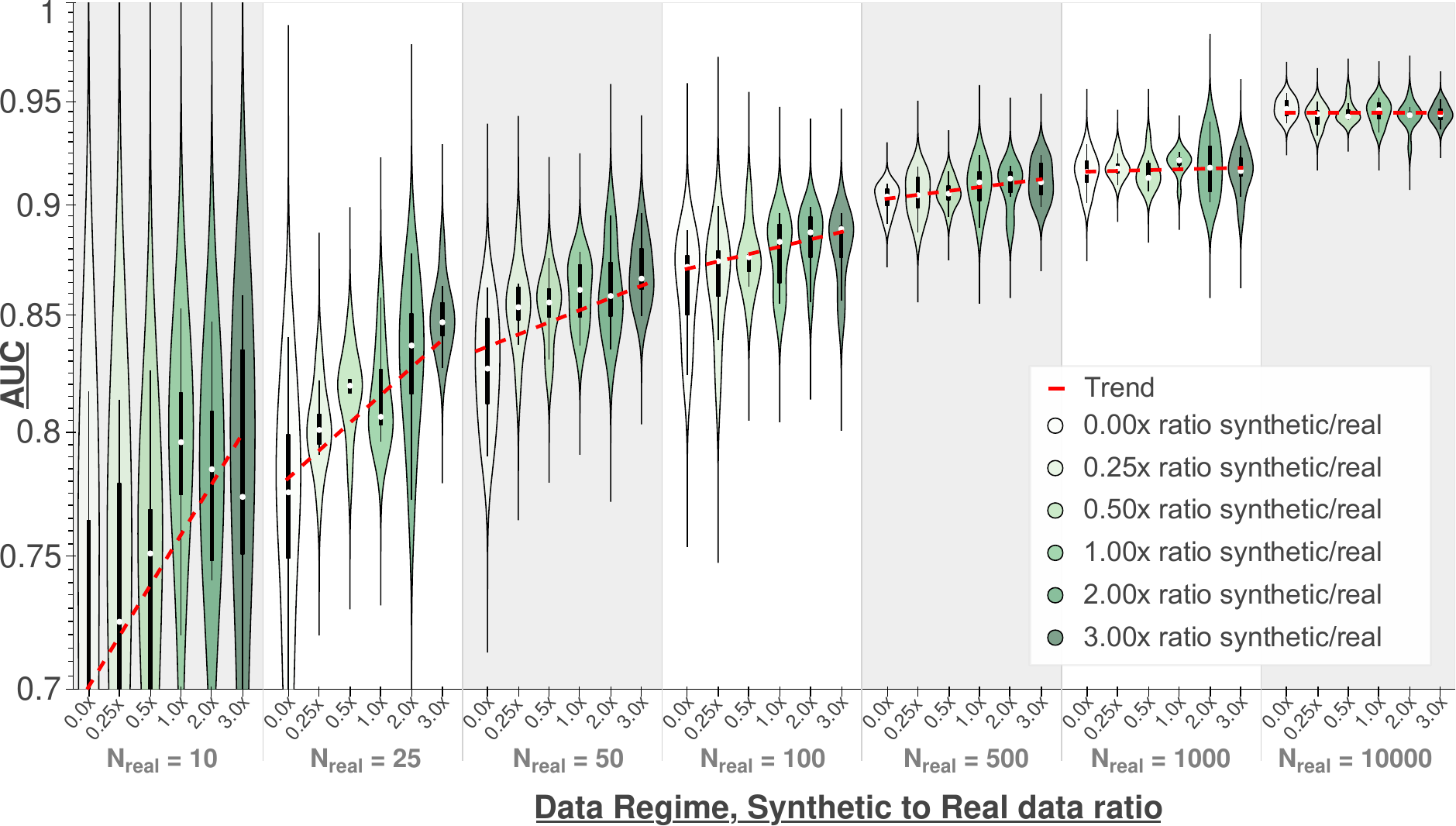}
    \caption{Distribution of test set AUC values across different folds for a classifier trained on varying proportions of real and synthetic data. Each group of violin plots represents a data regime with an initial real data size between 10 to 10,000 samples. Within each group, the multiple violin plots illustrate the test set AUC distribution when training the classifier with increasing amounts of synthetic data generated using our approach, at augmentation ratios of 25\%, 50\%, 100\%, 200\% and 300\% relative to the real data size (i.e. augmentation ratio of 200\% for data regime 100, corresponds to adding 200 synthetic samples to the 100 real sample training set). The increasing ratios are also visually described by the increasing green hue intensity. The violin plots depict both a box plot and kernel density plot, which allows a better visualization of the performance distribution of the 10 models trained for each setting. The median values are represented by a white dot.}
    
    \label{fig:multiple_dr}
\end{figure*}

First, a qualitative analysis was conducted to compare the quality of synthetic histology images generated by an out-of-the-box SD model with the ones obtained when fine-tuning SD class conditionally with the scarce available metadata (baseline approach). Results show that fine-tuning is an important step to ensure that SD learns to bridge the domain gap between the natural images it was pretrained with and the histological ones we aim to reproduce. The class-conditional fine-tuning allowed some degree of control on the presence cancerous and healthy features during image generation, which underlines SD's ability to represent complex biomedical concepts when captioned data is available.

Next, we studied the ability of the morphology-enriched method to improve upon the class-conditional baseline. Quantitative and qualitative image quality assessment reveals that the synthetic image fidelity and coverage of the real distribution increases when using the morphology-enriched methodology. In particular, this morphology-enriched prompting approach allowed a 88.6 decrease in FID score and an 0.11 increase in precision. However, the baseline synthetic dataset shows higher recall than the morphology-enriched method by 0.078, which is inconsistent with respect to the UMAP embedding visualized in Figure \ref{fig:coverage-with-samples-methods}. The UMAP embedding suggests that the morphology-enriched approach should have higher precision (which measures how much the morphology-enriched approach can be generated by the real morphology space) and the recall (which measures how much of the real morphology space can be generated by the morphology-enriched approach). We hypothesize four different explanations for this discrepancy. First, the UMAP projected embedding space might not precisely represent the original embedding space, complicating interpretation. Second, precision and recall have been computed from Inception-V3 embeddings which, although standard for their computation, are representative for natural images but might not perfectly represent in-domain histology images \cite{chambon2022roentgen}. Third, the morphology-enriched method might improve precision at the cost of ``\textit{non-realistic}'' diversity as quantified by the recall metric, thereby having a better trade-off between these two aspects \cite{sajjadi2018assessing}. Finally, given its very low precision (0.065), it is likely that this extra recall is due to the baseline set covering a large region of the Inception-v3 feature space (see Section \ref{sec:methods-feature_space}), going beyond the boundaries of the real data distribution, which is not desirable. 

In another test to address synthetic generation performance, a classifier was trained solely with synthetic data (Table \ref{tab:CAS-comparison}). This test also indicates the advantage of the proposed morphology-enriched prompt-building approach relative to the baseline: although training on only synthetic data is still not sufficient to reach the performance when training on only real data, the higher performance when training on images generated with the morphology-enriched method compared to the baseline method seems to confirm its improved ability to replicate the ``cancer'' and ``healthy'' tissue features present in the real dataset.

To address the visual quality of the generated samples, a visual Turing test was performed. The results of the visual Turing test suggest that distinguishing between real and synthetic examples was challenging, with poor sensitivity and specificity. However, two results do suggest that detectable differences may exist. Firstly, one of the readers achieved a statistically significant discriminative power between real and synthetic images. Nevertheless, this is likely to be explained by the presence of resampling artifacts in real images, rather than due to other meaningful image features. These artifacts likely originated from the downsampling of PCam's original WSIs from 40x magnification to 10x, in combination with the subsequent image upscaling for visualization purposes. Secondly, considering the slightly higher inter-reader agreement when the images were synthetic (see Figure \ref{fig:turing_reader_uncertainty_agreement_artifact}a), we cannot exclude that readers might have been more suspicious about synthetic images. Nevertheless, in general, synthetic image detection was not reliable.

Debriefing with pathologists generally confirmed the low reliability and confidence metrics measured. Readers reported that they would have exclusively chosen ``can not reliably distinguish'' instead of synthetic or real if they would have been given the choice, and that for many to most images the choice for synthetic or real was the result of a guess. In addition to the artifact discussed above, only few features had been tentatively assumed to be potential distinguishing features during the experiment (slightly increased cellular  monomorphism, order of tissue architecture, "rectangular" cell features, and nuclear hue). As analyzed later in the discussion, future work could further explore these and other potentially differentiating features.

Debriefing, on the other hand, also revealed four factors which may have relevantly decreased the potential sensitivity of readers to find ``flaws'' in synthetic images. Firstly, images were based on the PCam set, and thus corresponded to a small field of view and a low scanning resolution (see below). In addition, they were enlarged on the monitor for viewing. Pathologist reported that this made the images appear of much ``poorer quality'' than what they are used to from high-quality scans, and that this likely biased them towards assuming images cannot be differentiated and less thorough discrimination. Secondly, the tissue of the PCam set consists mostly of neoplastic cells, lymphocytes and stroma. Neoplastic cells however, can have an atypical morphology (cellular/nuclear pleomorphism, anisokaryosis/-cytosis, karyomegaly etc.), which makes it harder to decide whether an atypical morphology observed on an image is real but atypical, or a ``flaw'' in the synthetic image. (The remaining lymphocytes and stroma have comparatively little discerning features at the resolution used). For future work, it could be valuable to additionally explore ``more challenging'' scenarios, with highly ordered and feature-rich morphologies (e.g. non-neoplastic and higher resolution). Thirdly, as opposed to routine examination, readers were not allowed to navigate back and forth between images or directly compare images. Finally, there were time constraints due to balancing annotation with routine diagnostic work. 

All things considered, the results of the previously discussed image quality assessment, coupled with the challenge faced by expert pathologists in distinguishing between real and synthetic images, show that the generated images were impressively indiscernible from real images, confirming that their use as data for training AI models is plausible. Nevertheless, the visual Turing test only addressed whether a human expert was able to easily distinguish the images under the conditions used, neither ensuring that all specific features necessary to train a specific classifier are present, nor excluding potential detrimental (although minute) hallucinations which could be difficult to spot. 

To address these latter factors, the morphology-enriched method was further evaluated on its ability to produce synthetic images with downstream usability for improving the performance of a cancer CAD system. The results show that the high image quality and diversity produced by the proposed morphology-enriched method can be translated into actual practical utility, yielding performances comparable to using larger amounts of real data (as seen in Figure \ref{fig:multiple_dr}). While it is important to note that synthetic data cannot fully replace real data (as reported in Table \ref{tab:CAS-comparison}), these findings indicate that our proposed morphology-enriched method could be used for augmenting small training datasets in cases where only a few available slides exist that contain a morphology of interest, or cases where only simple patch-level annotations are available and do not come with additional metadata. Moreover, the approach could be used by data owners, who could generate synthetic data from their private datasets (even with limited annotations) to mitigate cost and privacy concerns related to their use. As an example, synthetic data could be shared jointly with small subsets of real private data for model training, thereby effectively lowering data acquisition costs. Consequently, training CAD systems could become more accessible, ultimately improving their medical utility.

In light of these findings, especially considering the image quality comparison between the two explored prompt-building approaches, there is a strong indication of the value of our approach when generating high-quality images from datasets with limited annotations. This is especially relevant in histopathology as the field lacks extensive and well-annotated datasets like those of other medical domains \cite{chambon_roentgen_2022, pinaya_brain_2022}. In Histopathology, annotations on the WSI level are much more readily available than annotations for specific regions, patches, or pixels, as routine pathological examination reports are usually recorded on the organ or case level. Moreover, if pixel-level annotations are required, their creation usually involve non-standard, time-consuming and costly processes. Therefore, considering the gigapixel scale of WSIs and the technical challenges associated to their processing, most datasets and AI models in this field are usually restricted to a collection of small patches with incomplete annotations, which, as shown here, limits the fidelity and realistic diversity of the images derived by these generative models. By circumventing the need for extensive annotations, our approach fills a critical gap in histopathology image generation, as it holds the potential to extend the advantages of synthetic data to a broader range of medically relevant applications where these are lacking.

The particular focus of this work is the improvement of the performance of cancer CAD systems with synthetic images generated in annotation-scarce scenarios, although the choice of this particular downstream application was enforced merely by the dataset in use (PCam). In fact, the presented framework is not linked to a specific tissue or set of classes, thereby being equally applicable to any other medically relevant disease classification task in the digital pathology domain that suffers from a lack of annotated data. Indeed, the positive results presented in this study should serve as motivation for further investigations into its utility across diverse scenarios. Moreover, even though the morphology-enriched prompt-building solution is particularly relevant in the digital pathology field due to the known constraints related to annotation availability, it could also be applied to other imaging domains where comprehensive annotations might be lacking, with minimal modifications in the text prompt template.

Nevertheless, our prompt-building approach presents some limitations that should be addressed in derivative works. Despite its efficacy, both prompt-building approaches rely on templates, which fall short in terms of prompt interpretability. This limitation arises from its reliance solely on the cluster indexes, which do not convey any explicit description of the shared image features that they represent. Adding to that, the templates lead to a rigid prompt set that might hinder the SD model's ability to understand free-form user input in other downstream applications \cite{ruan2023promptbased}. GPT-based models have been used to rephrase, summarize and simplify text in a number of applications \cite{lu2023napss, yellapragada2023pathldm, lyu2023translating}, with notable success in improving performance in other natural language processing tasks \cite{dai2023chataug}. Introducing greater semantic variability in our prompts is left as future work.

Furthermore, achieving a test set AUC of over 0.85 with just 100 real training samples from an initial dataset of over 200,000 samples (see Figure \ref{fig:multiple_dr}) suggests that the dataset and classification task are notably straightforward \cite{graham2020dense}. Therefore, our results on this task might underestimate the benefits of synthetic data in other, more challenging tasks. We leave this experimentation for future work too.

Additionally, the PCam dataset is limited to relatively small patches. Our method produced images at a theoretical resolution of 0.1823 $\mu m$/px and an area of 93.3 $\mu m^2$. However, given the downsampling originally performed in PCam, the overall information present in the image (i.e., detail of visible cellular structures) approximately corresponds the information content of a scan at 10x and 0.972 $\mu m$/px. In this sense, evaluating the proposed morphology-enriched method for the creation of larger and ``higher magnification'' images would be promising. As an example, synthetic images with a magnification corresponding to a ``standard'' 40x objective (e.g., 0.25 $\mu m$/px) or a larger field of view (e.g., 0.237 $\mu m^2$) could be generated \cite{patel2021contemporary, meuten2016mitotic}.

Finally, it would be interesting to explore whether using a feature extractor model that has been pretrained on histopathology images, rather than natural ones, would improve the overall performance of the morphology-enriched method, as using domain-specific feature extractors has been reported to outperform general-purpose ones \cite{azizi2022robust, chambon2022roentgen}.

In conclusion, in this work we take an important step towards histopathology patch generation with text-to-image diffusion models from a dataset without comprehensive metadata. Our method proved effective at addressing that limitation by obtaining meaningful conditions in an unsupervised manner from the data itself. These conditions were shown to enhance synthetic image quality and diversity in comparison to class-conditional image generation alone. Improved real-data performance of downstream classifiers trained on synthetic data suggests generated images may also replicate discriminative features between different classes more reliably. Despite some concerns related to the patch resolution, the blinded evaluation by expert pathologists further supported these conclusions as the distinction between real and synthetic images proved to be challenging. Additionally, we show that these synthetic images can be used jointly with small subsets of real data to minimize the amount of real data needed to train a model with limited to no loss in performance. All things considered, the work presented here has two main implications. First, it provides a promising solution for extending the benefits of synthetic data to a wider range of medical applications where extensively annotated data is scarce. And, second, the proposed morphology-enriched prompt-building approach shows promising potential for reducing the data acquisition costs related to training CAD systems, improving its accessibility and, consequently, their medical utility.

\section{Methods}

\subsection{Data and Preprocessing}
\label{sec:methods-data}

The dataset used in this work is PatchCamelyon (PCam) \cite{veeling_rotation_2018, ehteshami_bejnordi_diagnostic_2017, kaggle_challenge}, which is a common image classification benchmark consisting of 327,680 patches of 96x96 px in size, extracted from 400 histopathologic scans of sentinel lymph node sections from breast cancer patients. These scans were initially acquired at a ``40x'' (i.e., 40x objective, corresponding to a 400-fold magnification) but were undersampled \textit{a posteriori} to yield a resolution of 0.972 $\mu m$/px, corresponding to a scan at 10x \cite{veeling_rotation_2018, ehteshami_bejnordi_diagnostic_2017, kaggle_challenge}. The version of this dataset that we used is the curated one from Kaggle \cite{kaggle_challenge}, which removes all duplicated patches and comes with a default train/test split. The train split contains 220,025 labeled images, where the only provided annotation is a binary label indicating the presence or absence of metastatic tissue in the 32x32 center region. The official test comprises 57,486 images for which no labels are provided, therefore all the reported classifier performance evaluations required the submission of the predictions to Kaggle.

The dataset was further preprocessed to remove image patches lacking meaningful tissue characteristics that we want to be able to synthesize, i.e., no-foreground patches. For this purpose, the patches including only white background or extremely dark due to scanning artifacts were filtered based on their mean saturation (S) and value (V) from the HSV color space via thresholding. Additionally, patches lacking significant color variation were removed based on their minimum standard deviation on each of the HSV channels. Blurry images were also removed by thresholding the variance of the Laplacian operator. Finally, the number of enclosed shapes was used to remove images with a very low number of cell-like shapes. The above steps yielded a clean dataset of 216,868 image patches. Despite the preprocessing efforts, a percentage of patches in the original PCam dataset are mislabeled, as reported by an internal team of pathologists. Although this may cause noise during model training and inference, it should not alter the results given that this error is systematic for both the train and test sets, and that models are trained and tested on thousands of patches which would average out these labeling mistakes.

\subsection{Generative Architecture}
\label{sec:methods-modelarch}

The image generator architecture is a vanilla LDM model, fine-tuned from a specific pretrained model by the name of Stable Diffusion (SD)  \cite{rombach_high-resolution_2022}. It comprises 2 main components: a \textit{vector quantized} \textit{variational autoencoder} (VQ-VAE) and a \textit{conditional denoising UNet}.

The VAE is a stand-alone model that is pretrained to convert a high-dimensional input into a lower-dimensional representation from which the original data can be retrieved with minimal information loss. In the context of LDMs, the encoding branch of the VAE ($\mathcal{E}$) is used to encode the training images $x \in \mathbb{R}^{H \times W \times 3}$ onto a lower dimensional latent space $z = \mathcal{E}(x) \in \mathbb{R}^{h \times w \times c}$  (with  $h < H$ and $w < W$ ), upon which the training and generation process takes place. Here, $H$ and $W$ are the height and the width of the input images and $3$ represents the three color channels present in a typical WSI, while $h$, $w$ and $c$ are the, height, width and number of channels of the latent $z$. Latents in this space can be decoded back to the original image space via the VAE´s decoder branch $\hat{x} = \mathcal{D}(z)$.

In turn, the backbone of the LDM is a \textit{conditional denoising UNet} whose function is to model the VAE's latent space. In particular, this network is trained to minimize a denoising objective in that space via an iterative approach and while incorporating in that process optional conditional information. This conditional information can range from class labels, segmentation masks, to text, which is the one used in our pipeline. For the case of text, a CLIP model  \cite{radford_learning_nodate} is used to convert the text into rich embedded representations that can be introduced in the denoising process through a cross-attention mechanism.

Considering all these components, a training step of an LDM with text conditioning can be separated into two distinct processes, the \textit{forward} and \textit{reverse diffusion} steps. In the \textit{forward} process, given an image from the training data distribution ($x_0$) we compute its representation in the latent space of the VAE ($z_0 = \mathcal{E}(x_0)$) as well as the CLIP text embedding of its corresponding text prompt ($c(y)$). A time step is sampled from a Uniform distribution ($t \sim \mathcal{U}({1, ... , T})$) and random noisy latent is sampled from a Normal distribution ($\epsilon \sim N(0,1)$). A corrupted latent ($z_t$) is created by combining $\epsilon$ with $z_0$ via a scheduler that depends on the sampled $t$, ensuring that the larger the $t$ the more noise is added to $z_0$. In the \textit{reverse} diffusion process a UNet is used to predict the initially sampled noise, $\epsilon$, based on the sampled $t$, $z_t$ and its corresponding text embedding $c(y)$.

The MSE loss between the predicted and true noises allows the computation of a gradient that can be used to update the weights of the text encoder (CLIP) and, most importantly, of the UNet. See the Equation below:
\begin{equation}
    \textrm{L}_{LDM} = \textrm{E}_{(z \sim \epsilon(x), \, y, \, \epsilon \sim \mathcal{N} (0, 1), \, t} \left[ || \epsilon - \epsilon_{\theta}(z_t, \, t, \, c(y))||_2^2 \right]
\end{equation}
\label{eq:loss_ldm}
For image synthesis, a random noisy latent is sampled ($z_T \sim N(0,1)$) and using the user-provided conditioning text ($y$) it is possible to recursively denoise $z_T$ into a new uncorrupted latent ($z_0$). A new image can then be reconstructed with the decoding branch of the VAE, generating a new image $x=\mathcal{D}(z_0)$ which matches the description provided by $y$. In our case, given SD´s particular pretraining, the synthetic images are generated at a resolution of 512x512.

\subsection{Morphology-enriched prompt-building approach}
\label{sec:methods-pb}

As described earlier, the work here presented explores two different prompt-building approaches for fine-tuning SD. While the baseline approach leverages solely the provided patch labels for class-conditional image synthesis, our morphology-enriched approach further extracts semantic information from the images for more varied and realistic image generation.

As previously mentioned, the proposed morphology-enriched prompt-building approach is a two-step process. First, image features were extracted from each image in the dataset by a pretrained DINO-ViT \cite{caron_emerging_2021}, which has been shown to encode useful semantic information across tasks and domains \cite{caron_emerging_2021,amir_deep_2022}. The base ViT variant with 8x8 patches (B/8) was employed, which had the best linear classification performance in ImageNet \cite{caron_emerging_2021}. The output feature vector was set to be equal to the class token, a 768-dimensional vector that aggregates information from the entire image. Feature extraction was implemented using the Hugging Face Transformers library \cite{wolf-etal-2020-transformers}, and input images were preprocessed as suggested by the model creators. Next, these lower-dimensional representations of the histological images are grouped into several subgroups with shared morphological features, which we can later reference in the text prompts. To do so, the feature dataset was subject to unsupervised clustering via K-means, wherein the optimal \textit{k} parameter was selected by sweeping the \textit{k} values from 2 to 50 and selecting the one that allowed best combination of inter-cluster and intra-cluster variance as quantified by the SD index \cite{halkidi2000quality}. The optimal \textit{k} parameter was found to be at 33, yielding 33 different clusters of DiNO-ViT image features.

The template described in section \ref{sec:results-sd-proposed} is filled with both the existing patch labels and extracted cluster information for each image. Following this method, 66 unique text prompts (based on 2 labels and 33 clusters) were generated for the 216,868 image dataset which, compared to the \textit{baseline approach}, encode for a much wider variety of tissue morphologies and staining profiles, complementing the existing label information.

\subsection{Dataset Balancing}
\label{sec:methods-rs}

Due to the diversity and heterogeneity of histopathology data, it is likely that the distribution of the different tissue morphologies and staining profiles is not uniform across the curated PCam dataset. To prevent this unbalanced distribution from introducing bias during both the SD´s fine-tuning or inference towards specific image subtypes, we resampled the dataset to obtain a distribution of examples across prompts and classes that was as close as possible to uniform. The top 21 most populated prompts from each class (42 in total) were selected and undersampled to create a subset of 51,000 examples where every prompt and class label is approximately equally represented (either 1214 or 1215 examples per prompt). This subset is then split into a 50,000 training sample subset and 1,000 validation sample subset, also ensuring the same balanced prompt distribution is kept in both sets. The choice of the final curated balanced dataset´s size determines the number of prompts that can be selected. The 50,000 examples are used to minimize computational effort during the SD´s fine-tuning while also being large enough to provide statistical robustness for the downstream image quality metric computations. The 1,000 validation samples are used as an independent holdout set used to monitor the performance of the downstream classifier models during training so as to determine the best model configuration that should be used for testing, which is standard practice in Deep Learning.

Moreover, despite this curated subset having been selected based on the morphology-enriched prompt set, the same exact 50,000 training images are used for the baseline approach so as to ensure comparability between the two methods. This is achieved by duplicating the selected dataset and simply removing the extra morphology information from the prompts (i.e. \textit{``(...), morphology type $\left\langle\textnormal{\textbf{INDEX}}\right\rangle$''} ). Given the nature of both prompt-building approaches, this will also ensures that the balanced distribution across prompts and labels is maintained in the dataset captioned via the baseline approach.

\subsection{Fine-tuning and synthesis implementation details}
\label{sec:methods-ft_synthesis_details}

The same 50k curated dataset from PCam is prompted via the two different approaches explored in this work and then each set of prompts is used independently to fine-tune SD, yielding two different SD models. Both fine-tuning experiments were run for 12.5k training steps, using a float16 precision (fp16), a learning rate of $10^{-5}$ and with an effective batch size of 64 (4 GPUs with batch size of 8 and gradient accumulation of 2). The GPUs used were NVIDIA A10G Tensor Core GPUs and took approximately 12h for 12.5k training steps.

All models were retrieved from the HuggingFace model repository \cite{wolf2019huggingface}. The SD checkpoint version 1.5 was used (\textit{runwayml/stable-diffusion-v1-5} \cite{rombach_high-resolution_2022}) whereas the UNet was pretrained on LAION-2B (\textit{laion/laion2B-en} \cite{schuhmann2022laion}) and the VAE was pretrained on LAION-5B \cite{schuhmann2022laion}. The version of the CLIP model used for the conditioning mechanism was \textit{openai/clip-vit-large-patch14} \cite{radford_learning_nodate}, which was also trained on natural images. The SD pipeline and weights, downloaded from HuggingFace, were fine-tuned using \textit{text2image} in the \textsc{diffusers} library \cite{von-platen-etal-2022-diffusers}, while both the \textit{VAE} and CLIP models were frozen (i.e., no weight update). The \textsc{PyTorch 1.13.1} library is used as the underlying Deep Learning framework.

The resulting fine-tuned LDMs are then used for inference with the corresponding prompts, generating a synthetic dataset of 50k images also balanced on prompts and class, mirroring the corresponding real ones, but at an increased resolution of 512 x 512 px. The image generation pipeline for generating both these synthetic datasets was parameterized with a guidance scale of 7.5, 50 inference steps and a PNDM noise scheduler \cite{liu2022pseudo}.

\subsection{Metrics used to evaluate image quality and coverage}
\label{sec:methods-imqual}

The quality of the synthetic images was quantitatively measured via the Fréchet Inception Distance (FID) \cite{heusel2017gans}, which is the standard evaluation metric for assessing how close the distribution of real and synthetic data are. Rather than comparing the images pixel by pixel it based on the mean and standard deviation across each dataset of the corresponding Inception-V3 features (1x1028) \cite{szegedy_rethinking_2016}, a convolutional neural network pretrained on a large natural image dataset (ImageNet). The precision and recall \cite{kynkaanniemi2019improved} were two other metrics computed to provide independent assessments of average sample quality and coverage of the real distribution, respectively. Like the FID score, these last two metrics were also computed using Inception-V3 features. All metric computations were based on 50k samples from both the real and synthetic datasets, which is standard practice when reporting results of generative AI models. Finally, all metrics were computed using the PyTorch Image Quality (\textit{piq}) library \cite{kastryulin2022piq, piq}.

\subsection{UMAP feature space}
\label{sec:methods-feature_space}

UMAP is a recent dimensionality reduction technique that has been shown to project high dimensional data into useful lower dimensional layouts while preserving as much of the local and more of the global data structure than other methods, like t-SNE \cite{van2008visualizing}, with a shorter run time \cite{heusel2017gans, becht2019dimensionality}. The 2D UMAP \cite{mcinnes2018umap} projections of each Inception-v3 \cite{szegedy_rethinking_2016} feature vector were calculated and the overlap between the real and each synthetic distribution was visualized (see Figures \ref{fig:cancer_healthy_grids} and \ref{fig:coverage-with-samples-methods}). Notably, the 2D UMAP model was fit on only the real data and then used to project every real and synthetic latent to that learned space.

The regions A to F from the UMAP feature space highlighted in Figure \ref{fig:coverage-with-samples-methods} were manually selected to highlight the regions of the real data distribution from which the model fine-tuned with the baseline prompts cannot sample. Regions R\textsubscript{1} and R\textsubscript{2} were also manually selected, but represent specific subtypes of images from the real distribution that both explored approaches are able to replicate.

\subsection{Synthetic-only Classification: Experimental Details}

As described in Section \ref{sec:results-sd-proposed}, the validation of our proposed morphology-enriched approach is also supported by the training of a classifier on each of the following three sets of training data and their subsequent evaluation on the hidden test from Kaggle (see Section \ref{sec:methods-data}): (a) the 50k real dataset used to train both LDMs, (b) a 50k synthetic dataset generated via the baseline approach and (c) a 50k synthetic dataset generated via the proposed prompt-building approach. This was used to evaluate the extent to which synthetic data is able to represent the full distribution of real samples, and also to determine which of the methods better represents the discriminative features between the two classes in the real dataset.

The classifier architecture and hyperparameters chosen for this analysis were similar to those used in Yellapragada \textit{et al.} \cite{yellapragada2023pathldm}. It consisted of an ImageNet-pretrained ResNet-34 which is fine-tuned for 40 epochs with binary cross entropy loss with the Adam optimizer \cite{kingma2017adam} with default parameters. The learning rate was scheduled to be $10^{-3}$ for the first 20 epochs being reduced to a tenth of the current value at that point and also at the 30th epoch mark. No data augmentation transforms were used to address synthetic data performance, although the image patches were processed to meet the format and value range expected by the ImageNet weights. The best epoch was selected for testing based on its AUC score on the validation subset of 1000 real images described in \ref{sec:methods-rs}. Once more, the \textsc{PyTorch 1.13.1} library was also used as the underlying Deep Learning framework for this experiment.

\subsection{Visual Turing Test: Experimental Details}
\label{sec:methods-visual_turing}

\begin{table}[t!]
    \centering
    {
    \scriptsize
    \begin{tabular}{llll}
    \begin{tabular}[c]{@{}l@{}}\textbf{Kappa} \\ \textbf{values}\end{tabular}   & \begin{tabular}[c]{@{}l@{}}\textbf{Interpretation of} \\ \textbf{level of agreement}\end{tabular} & \begin{tabular}[c]{@{}l@{}}\textbf{Kappa} \\ \textbf{values}\end{tabular}                                                & \begin{tabular}[c]{@{}l@{}}\textbf{Interpretation of}\\ \textbf{Level of agreement}\end{tabular}                                        \\ \hline\hline
    \rowcolor[HTML]{EFEFEF} 
    1.00           & Perfect agreement                                                               & \cellcolor[HTML]{EFEFEF}                                    & \cellcolor[HTML]{EFEFEF}                                                                                              \\
    \rowcolor[HTML]{EFEFEF} 
    0.93-0.99    & Excellent agreement                                                             & \multirow{-2}{*}{\cellcolor[HTML]{EFEFEF}\textgreater 0.75} & \multirow{-2}{*}{\cellcolor[HTML]{EFEFEF}\begin{tabular}[c]{@{}l@{}}Excellent agreement\\ beyond chance\end{tabular}} \\
    0.81-0.92    & Very good agreement                                                             &                                                             &                                                                                                                       \\
    0.61-0.80    & Good agreement                                                                  & \multirow{-2}{*}{0.4-0.75}                                & \multirow{-2}{*}{\begin{tabular}[c]{@{}l@{}}Very good agreement\\ beyond chance\end{tabular}}                         \\
    \rowcolor[HTML]{EFEFEF} 
    0.41-0.60    & Substantial agreement                                                      & \cellcolor[HTML]{EFEFEF}                                    & \cellcolor[HTML]{EFEFEF}                                                                                              \\
    \rowcolor[HTML]{EFEFEF} 
    0.21-0.40    & Slight agreement                                                                & \cellcolor[HTML]{EFEFEF}                                    & \cellcolor[HTML]{EFEFEF}                                                                                              \\
    \rowcolor[HTML]{EFEFEF} 
    0.01-0.20    & Poor/chance agreement                                                           & \multirow{-3}{*}{\cellcolor[HTML]{EFEFEF}\textless 0.40}    & \multirow{-3}{*}{\cellcolor[HTML]{EFEFEF}\begin{tabular}[c]{@{}l@{}}Poor agreement\\ beyond chance\end{tabular}}      \\
    $\leq$ 0 & No agreement                                                                    & \multicolumn{1}{c}{-}                                       & \multicolumn{1}{c}{-}                                                                                                 \\ \hline
    \end{tabular}
    }
   \caption{Qualitative interpretation of the Cohen’s Kappa coefficients as proposed by \cite{dawson2004, landis1977}.}
   \label{tab:interpretation-kappa}
\end{table}

A set of 20 real and 20 synthetic images, generated by the morphology-enriched prompt-building approach, were presented to a group of five board-certified veterinary pathologists, using the Ground Truth suite within AWS SageMaker (Image Classification, Single Label). The number of examples was set to only 40 samples to minimize the labeling effort required by the readers, while still enabling sufficient statistical power. To ensure that the main subtypes of images from the real and synthetic dataset are well represented in this small subset, the images were selected to cover multiple regions of the 2D UMAP space. Real Images were upscaled from 96x96 px to 512x512 px to match the synthetic images, and displayed on pathologists' screens at a resolution of approximately 96 dpi.

Each reader was tasked with classifying each image as either “definitely real”, “maybe real”, “maybe synthetic” or “definitely synthetic”. Notably, a neutral option was deliberately omitted from the response choices. This experimental design allowed for dichotomized responses, while concurrently capturing readers' confidence. Images were presented sequentially and reader responses were recorded before progressing to subsequent images. To limit transductive reasoning, readers were not allowed to skip, revisit, or navigate back and forth between images, i.e., each image could only be viewed once. Furthermore, readers worked independently, had not been exposed to examples of real or synthetic images before the evaluation task, and were not given further background information on the goal of the project. Readers were asked to perform a relatively quick evaluation not impinging on their daily routine work and the time to evaluate the images was otherwise left at the readers discretion. Finally, readers could provide comments along with their answers, which were recorded for later analysis.

The readers’ labeling performance was evaluated using accuracy, sensitivity and specificity against our ground truth. We used a two-sided binomial test for the null hypothesis that readers cannot discriminate between real and synthetic images (expected accuracy = 0.5). Furthermore, we estimated the readers’ labeling confidence both subjectively and objectively. The former was evaluated based on the percentage of their high-confidence (qualified by ``definitely'') and low-confidence (qualified by ``maybe'') labels. The latter was measured using the lead time (time taken to answer), assuming that a higher lead time could indicate higher label uncertainty (see Section \ref{sec:extra-leadtimes} of the supplementary material). Inter-reader reliability (IRR) was determined pairwise between independent readers and was estimated for each presented image using Cohen’s Kappa statistic \cite{mchugh2012interrater}. Cohen’s Kappa coefficients can range from -1 to 1, where 1 indicates perfect agreement, 0 indicates agreement expected by chance and -1 indicates complete disagreement between readers. The somewhat arbitrary value of 0.21 was considered a minimal reliability criterion as suggested by the literature \cite{dawson2004, landis1977}. Table \ref{tab:interpretation-kappa} describes how the kappa values should be qualitatively interpreted according to the same sources.

\subsection{Synthetic Data Augmentation: Experimental Details}
\label{sec:methods-synth_aug}

As described in \ref{sec:results-synth_aug}, the evaluation experiments relied on using multiple proportions of real and synthetic data to train classifiers for distinguishing ``cancer'' labeled patches from ``healthy'' ones. In particular, for each synthetic dataset, we conducted experiments of adding synthetic data in different proportions in 7 different training data regimes. Each data regime corresponds to an initial amount of real data and spanned over the following range of values:  10, 25, 50, 100, 500, 1000, 10000 real samples. For each data regime, we then added synthetic samples to the training data in 6 different ratios with respect to the initial amount of real data: 0\%, 25\%, 50\%, 100\%, 200\%, 300\%. See Figure \ref{fig:multiple_dr}.

A K-fold cross-validation scheme was used to ensure the robustness of our conclusions to the subsets of real and synthetic data used for training the classifiers. For each initial amount of real data and for each synthetic augmentation ratios, $k$ different models (in this case $k=10$) are trained on distinct subsets of real and synthetic data sampled using 10 different fixed random seeds. In total 420 different models were trained across all training settings (7 data regimes, 6 ratios, 10 trainings lead to  $420 = 7 \times 6 \times 10$ trained models). The results are then reported as a distribution of the Kaggle test set performances obtained by each of the 10 models for each setting.

The architecture chosen for the classifier consisted of a ResNet-50, initialized from a checkpoint pretrained on ImageNet. No data augmentation transforms were used, but the patches were processed to meet the format and value range expected by the model. The training ran until convergence with early stopping with patience of 20 epochs for 70 epochs at max, while monitoring the validation loss on the validation set of 1000 images described earlier in section \ref{sec:methods-rs}. A batch size of 8 was kept the same for all data regimes for consistency. The loss used was the binary cross entropy with the Adam optimizer \cite{kingma2017adam} with default parameters and a learning rate of $10^{-4}$. Once again, the \textsc{PyTorch 1.13.1} library was also used as the underlying Deep Learning framework for this experiment.


\section*{Data availability}

PatchCamelyon, the dataset used for this work, is a publicly available dataset comprising 400 WSI images of lymph node sections. No external data sources were used for training or evaluating the models. According to the authors \cite{veeling_rotation_2018, ehteshami_bejnordi_diagnostic_2017}, the slides were ``\textit{acquired and digitized at 2 different centers using a 40x objective (resultant pixel resolution of 0.243 microns)} [and they] \textit{undersample this at 10x to increase the field of view}''.

\printbibliography

\section*{Acknowledgments}

We would like to thank Jeanne Kehren for sponsoring the activity and her continued support, as well as to  Marion Legler and Hendrik Esch for their support, constructive conversations and inspiration.

\section*{Author contributions}

P.O., G.DB., S.Si. and S.M. conceived the presented idea and formulated the research goals. P.O. and S.Si. devised and implemented the data preprocessing. P.O., G.DB. and G.JP. wrote the code for conducting the prompt-building as well as the training of the SD model and all the evaluated classifier models. P.O., G.JP., G.DB., J.M.T., J.H. and S.M. designed the experiments. P.O. and G.JP. conducted and analyzed the results from the image quality assessment as well as all the classifier experiments. J.M.T., J.H., M.R., U.B, S.Sc., K.S., J.V., B.L. and G.JP. contributed to designing, preparing the code, executing and analyzing the results from the reader study. P.O., G.JP., J.M.T., J.H., S.M., M.R. and S.Si. contributed to the preparation of the manuscript. S.M. contributed to design the project.

\onecolumn


\section{Supplementary Material}

\subsection{Lead Time Analysis}\label{sec:extra-leadtimes}

\begin{table*}[h!]
\centering
\begin{tabular}{c|ccccc}
\hline
Reader Id & \begin{tabular}[c]{@{}c@{}}Lead-time\textsubscript{Real} \\ ($\mu \pm \sigma$ sec)\end{tabular} & \begin{tabular}[c]{@{}c@{}}Lead-time\textsubscript{Synthetic} \\ ($\mu \pm \sigma$ sec)\end{tabular} & \begin{tabular}[c]{@{}c@{}}Intra-reader\textsubscript{Real vs Sythetic}\\ p-value\end{tabular} & \begin{tabular}[c]{@{}c@{}}Inter-reader\textsubscript{Real}\\ p-value\end{tabular} & \begin{tabular}[c]{@{}c@{}}Inter-reader\textsubscript{Synthetic}\\ p-value\end{tabular} \\ \hline
1         & $\text{13.82} \pm \text{12.29}$                                                                               & $\text{12.56} \pm \text{3.87}$                                                                                     & 0.15520                                                                                        & 0.260                                                                              & 0.289                                                                                   \\
2         & $\text{32.13} \pm \text{41.18}$                                                                              & $\text{54.47} \pm \text{108.94}$                                                                                  & 0.47304                                                                                        & \textbf{0.011}                                                                     & \textbf{0.042}                                                                          \\
3         & $\text{5.46} \pm \text{2.82}$                                                                                  & $\text{5.36} \pm \text{1.9}$                                                                                       & 0.49460                                                                                        & \textbf{0.000}                                                                     & \textbf{0.000}                                                                          \\
4         & $\text{16.87} \pm \text{7.94}$                                                                                & $\text{23.11} \pm \text{23.10}$                                                                                     & 0.09029                                                                              & \textbf{0.001}                                                                     & \textbf{0.001}                                                                          \\
5         & $\text{41.77} \pm \text{118.53}$                                                                               & $\text{27.49} \pm \text{40.18}$                                                                                      & 0.35749                                                                                        & 0.423                                                                              & 0.380                                                                                   \\ \hline
\end{tabular}
\caption{Results of the lead time analysis. The mean and standard deviation of the lead times according to the nature of the test image are described in the first 2 columns. The Intra-reader\textsubscript{Real vs Synthetic} p-values measure, for each reader, the statistical significance of the differences between the lead times when the test images were real and the lead times when the test images were synthetic. The following two columns (Inter-reader\textsubscript{Real} and Inter-reader\textsubscript{Synthetic}) depict the p-values measuring how significant are the differences in the time spent labeling the images between a given reader and all other readers, according to the nature of the test image. All these p-values were computed using a non parametric significance test (Wilcoxon Rank-sum test), to account for the non-gaussian distribution of these lead times (according to the Kolmogorov-Smirnov test). The p-values under the significance threshold of 0.05 are highlighted in bold.}
\label{tab:extra-leadtimes}
\end{table*}

Lead times (time to answer) were also measured and interpreted as a proxy for reader confidence. Table \ref{tab:extra-leadtimes} shows that, on average, readers spent $22.01 \pm 58.01$ seconds on real images and $24.60 \pm 55.59$ seconds on synthetic images. Nevertheless, no reader showed a statistically significant difference between the lead times depending on the nature of the test image (real or synthetic). We observed significantly lower lead times for reader 3 (real: $5.46 \pm 2.82$, $p < 0.001$, synthetic: $5.36 \pm 1.90$, $p < 0.001$) and reader 4 (real: $16.87 \pm 7.94$, $p < 0.001$, synthetic: $23.11 \pm 23.10$, $p < 0.001$, which suggests a higher confidence of these readers labeling decision, compared to other readers. Significantly higher lead times were observed for reader 2 (real: $32.13 \pm 41.18$, $p = 0.011$,  synthetic: $54.47 \pm 108.94$, $p = 0.042$). The high standard deviations for readers 2 and 5 can be explained to some extend by high lead time outliers ($> 500$ seconds).
\end{document}